\documentclass[letterpaper, 10 pt, conference]{ieeeconf}  

\IEEEoverridecommandlockouts                              

\usepackage{graphicx}
\usepackage{float}
\usepackage{amsmath}
\usepackage{subcaption}
\usepackage{amssymb}
\usepackage{comment}
\title{\LARGE \bf
Cognitive Energy Modeling for Neuroadaptive Human-Machine Systems using EEG and WGAN-GP
}

\author{
Sriram Sattiraju$^{1,*}$,
Vaibhav Gollapalli$^{2,*}$,
Aryan Shah$^{2}$,
and Timothy McMahan$^{2}$%
\thanks{$^{1}$Sriram Sattiraju is with the Department of Computer Science,
University of Texas at Austin, Austin, TX 78712, USA
(e-mail: srirams@cs.utexas.edu).}%
\thanks{$^{2}$Vaibhav Gollapalli, Aryan Shah, and Timothy McMahan are with
the Department of Learning Technologies, University of North Texas,
Denton, TX 76203, USA
(e-mail: VaibhavGollapalli@my.unt.edu;
AryanShah2@my.unt.edu;
Fred.McMahan@unt.edu).}%
\thanks{$^{*}$Sriram Sattiraju and Vaibhav Gollapalli contributed equally to this work.}%
}

\begin{document}

\maketitle
\thispagestyle{empty}
\pagestyle{empty}

\begin{abstract}

Electroencephalography (EEG) provides a non-invasive insight into the brain's cognitive and emotional dynamics. However, modeling how these states evolve in real time and quantifying the energy required for such transitions remains a major challenge. The Schrödinger Bridge Problem (SBP) offers a principled probabilistic framework to model the most efficient evolution between the brain states, interpreted as a measure of cognitive energy cost. While generative models such as GANs have been widely used to augment EEG data, it remains unclear whether synthetic EEG preserves the underlying dynamical structure required for transition-based analysis. In this work, we address this gap by using SBP-derived transport cost as a metric to evaluate whether GAN-generated EEG retains the distributional geometry necessary for energy-based modeling of cognitive state transitions. We compare transition energies derived from real and synthetic EEG collected during Stroop tasks and demonstrate group-level agreement and partial preservation of participant-level transition trends. These results indicate that synthetic EEG preserves the transition structure required for SBP-based modeling, enabling its use in data-efficient neuroadaptive systems. We further present a framework in which SBP-derived cognitive energy serves as a control signal for adaptive human-machine systems, supporting real-time adjustment of system behavior in response to user cognitive and affective state.

\end{abstract}

\section{INTRODUCTION}
EEG provides a non-invasive window into neural activity, capturing dynamic oscillatory patterns that reflect cognitive and emotional states in real time.  However, quantifying how these neural states evolve, particularly in terms of cognitive energy, remains an unsolved challenge. Cognitive processes such as attention, conflict monitoring, and emotional regulation are inherently dynamic, involving continuous state transitions that cannot be fully represented through static EEG features. Traditional methods for emotion or workload detection rely on static spectral power or connectivity measures, which fail to capture the probabilistic nature of brain-state evolution.

The Schrödinger Bridge Problem (SBP) provides a rigorous mathematical framework to quantify the minimum energy cost required for one probability distribution, representing a brain state, to evolve into another under stochastic dynamics. Originally developed in control theory and optimal transport, SBP has recently been validated in neuroimaging as a novel approach to measure brain state transition costs during cognitive tasks \cite{Kawakita2022}. This method allows for interpreting neural adaptation and mental effort as an energy optimization problem, aligning well with the concept of cognitive effort in EEG microstate research \cite{Barzon2024}. While SBP provides a principled framework for modeling cognitive state transitions, its application in data-limited settings remains challenging due to the scarcity of large-scale EEG datasets. This is where generative models, such as GANs, come in, being widely used to augment EEG data and demonstrating strong performance in preserving spectral and statistical characteristics \cite{Bouallegue2020}\cite{Lashgari2020}. However, existing validation approaches primarily focus on marginal distributions or downstream classification accuracy, leaving open the question of whether synthetic EEG preserves the underlying dynamical structure required for modeling state transitions.

In this work, we investigate whether GAN-generated EEG preserves the distributional geometry required for energy-based modeling of cognitive state transitions. Specifically, we use the Schrödinger Bridge Problem (SBP) as a principled metric of cognitive transport cost to compare transition energies derived from real and synthetic EEG collected during Stroop tasks. By evaluating agreement between SBP-derived energies across real and generated data, we assess whether synthetic EEG retains the structure necessary for modeling task-dependent neural adaptation. Building on this validation, we present a framework in which SBP-derived cognitive energy serves as a control signal for neuroadaptive human–machine systems. In this framework, EEG-derived brain state distributions are continuously evaluated to estimate cognitive effort, enabling real-time adjustment of system behavior in response to user state. Such a formulation supports adaptive environments, including gameplay and interactive interfaces, that dynamically regulate difficulty, pacing, or feedback based on cognitive load and affective state. By demonstrating that synthetic EEG preserves SBP-based transport structure, this approach also enables scalable training and deployment of neuroadaptive systems in data-limited settings.

\section{RELATED PRIOR RESEARCH}

A substantial body of work has examined synthetic EEG generation as a solution to limited data availability within Neural Modeling and Brain Computer Interface (BCI) research. Early studies demonstrated that GANs can learn complex temporal and spatial EEG distributions and produce synthetic signals suitable for downstream decoding tasks \cite{8834503}. These foundational efforts established that adversarial learning can model nonstationary neural signals beyond simple noise based augmentation.
    
GAN-based EEG synthesis was further extended to a variety of paradigms, such as motor imagery, and RSVP tasks. Several studies have demonstrated that when synthetic data is added to training pipelines, Wasserstein-based GAN (WGAN) architectures are very good at stabilizing training and maintaining event-related dynamics, which improves classification accuracy \cite{9130768}. These findings suggest that synthetic EEG can retain a cognitively meaningful temporal structure rather than simply approximating spectral statistics.
    
GAN-based augmentation has also been extensively applied to affective and emotion-related EEG analysis. Several studies report that synthetic EEG improves emotion recognition performance by increasing data diversity and reducing overfitting in deep learning models \cite{8512865}\cite{Liu2023EEGDA}. Conditional GAN formulations are particularly relevant, enabling class-consistent signal generation aligned with emotional or cognitive labels and supporting more controlled evaluation of model behavior.
    
In order to increase the physiological realism of synthetic EEG, more recent research has focused on domain-specific limitations. Alpha, Theta, and Beta EEG activity can be selectively maintained in artificial signals by conditioning GANs on spectral features linked to cognitive processes using frequency-aware generating techniques. The validity of limited generation procedures is further supported by related clinical approaches that demonstrate Conditional WGAN's ability to provide diagnostically consistent EEG across illness groups \cite{mutlu2025syntheticalseegdataaugmentation}\cite{Carrle2023-wf}. These studies support the idea that synthetic EEG can encode meaningful neural signatures beyond task-specific decoding.
    
While pointing out unsolved issues, a number of surveys and toolkits demonstrate the increasing maturity of GAN-based EEG modeling \cite{Barzon2024}\cite{Lashgari2020}. Reviews highlight that the majority of earlier research assesses synthetic EEG using classifier performance rather than more in-depth dynamical or physiological criteria. Recent frameworks aim to standardize EEG-focused GAN architectures and improve reproducibility across tasks and datasets \cite{Song2024-ym}.
    
Lastly, validation studies of consumer-grade devices have progressively indicated the viability of implementing sophisticated EEG modeling pipelines outside of lab settings. Despite increased noise and fewer channels, wearable EEG systems can record signals appropriate for cognitive and affective analysis, according to empirical assessments and systematic reviews \cite{Rieiro2019-ez}. This trend motivates investigating whether synthetic data and energy-based metrics remain reliable under realistic acquisition constraints.

Taken together, prior work demonstrates the effectiveness of GANs for EEG data augmentation and the promise of transition-based metrics for modeling cognitive effort, but it leaves open the question of whether synthetic EEG preserves the underlying neural dynamics required for energy-based state transition analysis. Addressing this gap is essential for validating synthetic EEG as a substitute for real recordings in adaptive and real-time neurotechnology systems.

\section{METHODOLOGY}

EEG data were collected from 10 participants between 18 and 24 years of age using an EMOTIV EPOC X wireless EEG headset during a multi-stage Stroop task designed to elicit progressively increasing levels of cognitive control and interference. The EPOC X records from 14 channels with CMS/DRL reference sensors. EEG signals were sampled at 256 Hz.

Each participant completed four tasks, corresponding to increasing cognitive demand. For each EEG segment, a set of features was extracted to represent instantaneous brain states in a multivariate feature space. These feature vectors define empirical probability distributions over neural states for each participant and task condition. To enable meaningful distributional comparison across participants, subject-wise baseline normalization was applied. For each participant, the first Stroop portion was treated as a baseline reference. Given a raw EEG feature vector
$x$, baseline-referenced normalization was performed as:
\begin{equation}
x' = \frac{x - \mu_{\text{baseline}}}{\sigma_{\text{baseline}} + \epsilon}
\end{equation}
where $\mu_{\text{baseline}}$ and $\sigma_{\text{baseline}}$ denote the participant-specific mean and standard deviation computed from baseline samples, and $\epsilon$ is a small constant added for numerical stability. This normalization removes inter-subject amplitude differences while preserving within-subject deviations associated with task-induced cognitive changes. Since the Schr\"{o}dinger Bridge Problem  (SBP) operates on probability distributions, this step is crucial to ensure that energy estimates reflect cognitive transitions rather than subject specific scaling. To further stabilize adversarial training and limit the influence of outliers, normalized features were clipped to a fixed range. All subsequent modeling and SBP analysis were performed in this baseline-referenced feature space. 
    
To address limited EEG sample availability and to model task-conditioned EEG distributions, a conditional Generative Adversarial Network (GAN) was employed. The generator maps samples from a latent noise distribution to EEG feature vectors while conditioning on both participant identity and task portion via learned embeddings. These embeddings are concatenated with the latent input, enforcing subject and task-consistent generation. The generator architecture consists of fully connected layers with residual connections, enabling stable learning in high-dimensional feature space.
    
The discriminator was implemented as a Wasserstein critic using a packed GAN (PacGAN) formulation. In this approach, multiple samples are concatenated and evaluated jointly by the critic, which reduces mode collapse and improves coverage of multimodal distributions. This design choice is particularly important for EEG data, where realistic modeling requires preserving the overall geometry of the distribution rather than matching marginal statistics alone. Training followed the Wasserstein GAN with gradient penalty (WGAN-GP) framework, which enforces Lipschitz continuity and improves convergence stability. The discriminator loss is given by:
    \begin{equation}
\begin{aligned}
\mathcal{L}_D
&= \mathbb{E}\left[D(\tilde{x})\right]
 - \mathbb{E}\left[D(x)\right] \\
&\quad + \lambda_{\text{GP}} \,
\mathbb{E}\left[
\left(
\left\|\nabla_{\hat{x}} D(\hat{x})\right\|_2 - 1
\right)^2
\right]
\end{aligned}
\end{equation}
    where $\tilde{x}$ denotes generated samples and $\hat{x}$ represents interpolated samples between real and generated data. 
    
While adversarial training aligns real and synthetic EEG distributions globally, SBP-based energy computation is sensitive to distributional geometry, particularly second-order statistics. To preserve variance structure within each participant and task condition, a condition-wise variance matching regularization term was introduced. For each participant–task group present within a training batch, per-feature variances of real and generated samples were computed and penalized when they diverged. The variance matching loss is defined as:
    \begin{equation}
    \mathcal{L}_{\text{var}}
    =
    \frac{1}{|\mathcal{G}|}
    \sum_{g \in \mathcal{G}}
    \frac{1}{d}
    \sum_{i=1}^{d}
    \left(
    \operatorname{Var}_{\text{real}}^{(g,i)}
    -
    \operatorname{Var}_{\text{gen}}^{(g,i)}
    \right)^2
    \end{equation}
    
    where ${|\mathcal{G}|}$ denotes the set of participant–task condition groups and $d$ is the feature dimension. The generator objective combines the adversarial loss and the variance regularization term:
    \begin{equation}
    \mathcal{L}_G
    =
    -
    \mathbb{E}\left[D(\tilde{x})\right]
    +
    \lambda_{\text{var}} \, \mathcal{L}_{\text{var}}
    \end{equation}
    
This regularization discourages artificial variance collapse and ensures that synthetic EEG distributions preserve covariance structure relevant for energy-based analysis. 
    
Synthetic EEG samples are generated in the baseline-referenced normalized space. For interpretability, samples can be transformed back into original EEG units using the inverse of the subject-specific normalization. However, all distributional comparisons and  Schr\"{o}dinger Bridge computations were performed in the normalized space to ensure consistency between real and synthetic data.
    
The  Schr\"{o}dinger Bridge Problem was used to quantify the energy cost associated with transitions between EEG-defined brain states. For each participant, empirical distributions $P_0$ and $P_1$ were constructed over the EEG feature space for two task portions. SBP computes the minimum stochastic control cost required to transport $P_0$ to $P_1$ under entropy-regularized stochastic dynamics. This cost is interpreted as the cognitive energy required for neural adaptation between task conditions.
    
The SBP formulation and numerical procedure were applied identically to distributions derived from real EEG data and from GAN-generated synthetic data. Agreement between SBP-derived transition energies across real and synthetic datasets serves as validation that the synthetic EEG preserves the distributional structure necessary for meaningful cognitive energy estimation and downstream neuroadaptive applications.

\section{RESULTS}

    This section presents the experimental results evaluating the fidelity of GAN-generated EEG data and its ability to preserve task-dependent cognitive transport structure as measured by Schrödinger Bridge Problem (SBP) energy. We analyze GAN training stability, feature-level distribution alignment, and both group-level and participant-level SBP transport costs, comparing real EEG recordings to GAN-synthesized EEG.
    \subsection{GAN Training Stability}
    
\begin{figure}[t]
        \centering
        
        \begin{subfigure}{\columnwidth}
        \includegraphics[width=0.98\columnwidth]{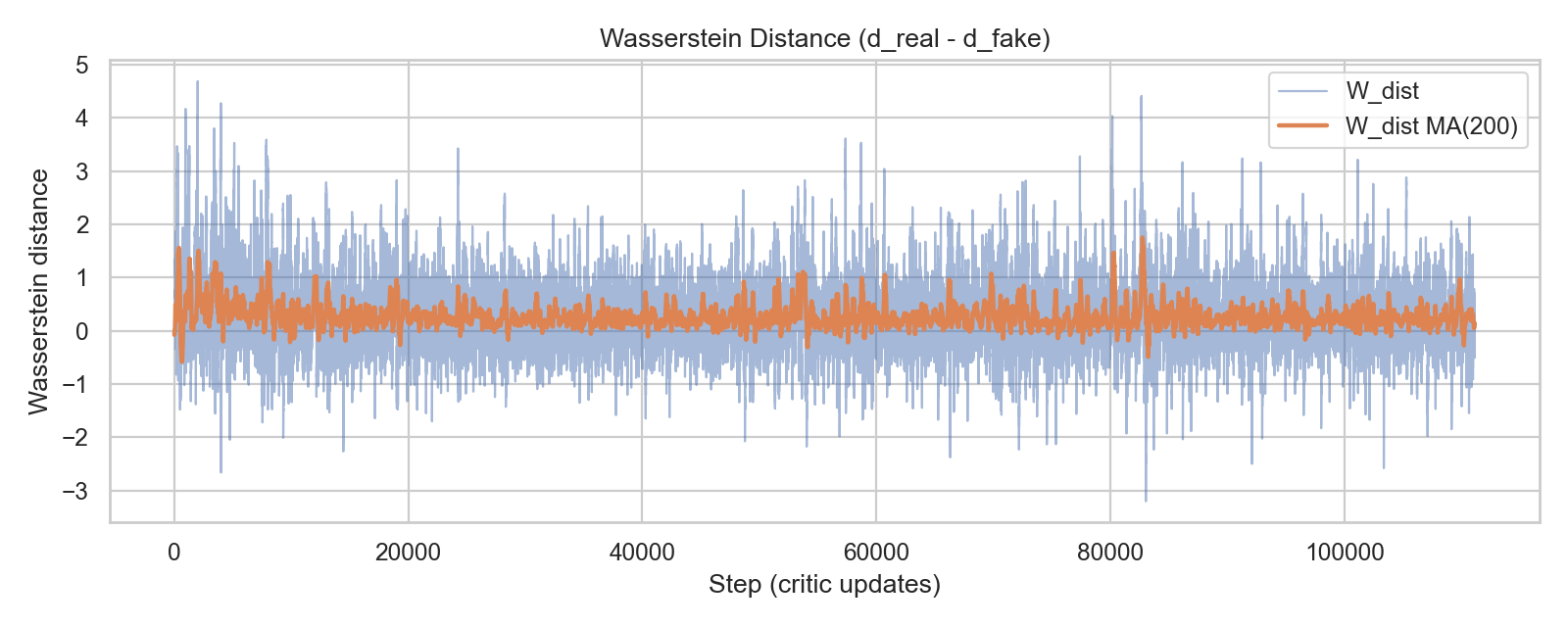}
        \caption{Wasserstein Distance vs. Critic Updates}
        \end{subfigure}
        
        \vspace{2mm}
        
        \begin{subfigure}{\columnwidth}
        \includegraphics[width=0.98\columnwidth]{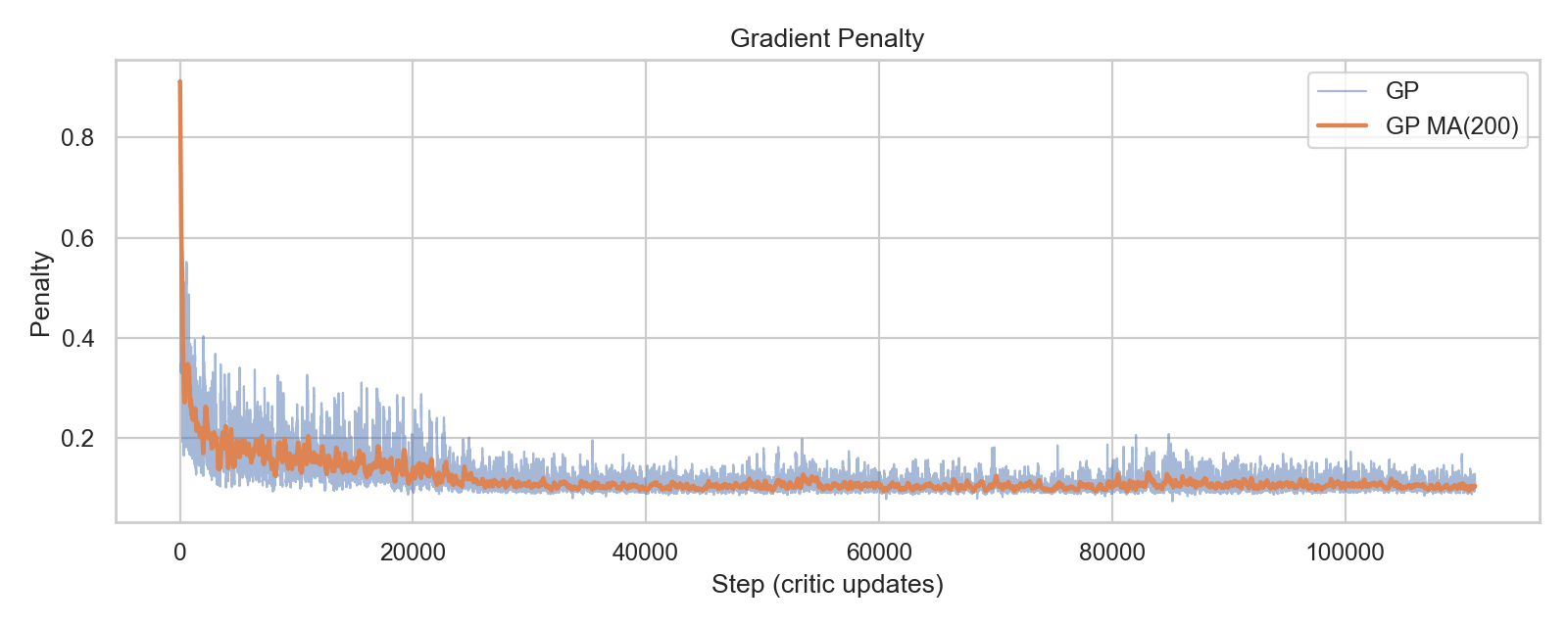}
        \caption{Gradient Penalty vs. Critic Updates}
        \end{subfigure}
        
        \caption{GAN training stability showing convergence of distribution alignment and bounded gradient penalty.}
        \label{fig:wasserstein_distance}
    \end{figure}

    To ensure that downstream Schrödinger Bridge Problem (SBP) analysis is performed on a reliable generative model, we first evaluate the stability of GAN training. Figure~\ref{fig:wasserstein_distance}(a) illustrates the evolution of the Wasserstein distance between real and generated EEG distributions across critic update steps, together with a moving average. After an initial transient phase, the Wasserstein distance stabilizes near zero without sustained divergence, indicating convergence of the generator--critic optimization.
    
    Figure \ref{fig:wasserstein_distance}(b) shows the corresponding gradient penalty throughout training. The gradient penalty rapidly decreases during early training and remains bounded for the remainder of the optimization, confirming that the Lipschitz constraint is successfully enforced. Together, these results demonstrate stable GAN training without evidence of mode collapse or numerical instability, providing a reliable foundation for subsequent SBP-based transport analysis.

    \subsection{Feature-Level Distribution Alignment}
    
\begin{figure*}[t]
        \centering
        \includegraphics[width=\textwidth]{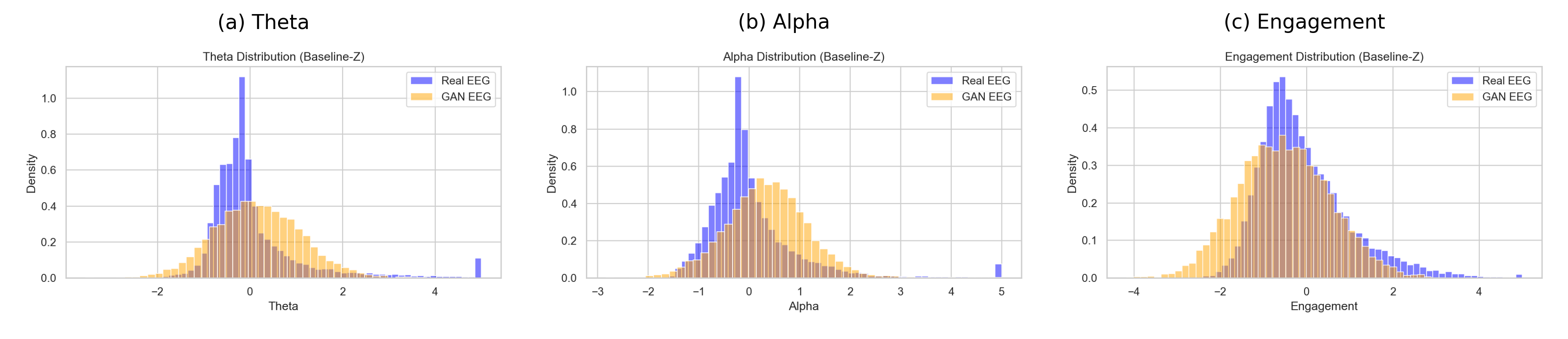}
        \caption{Comparison of real and GAN-generated EEG feature distributions for Theta power, Alpha power, and Engagement, showing alignment in distribution shape and variance.}
        \label{fig:3}
    \end{figure*}

    We next assess whether GAN-generated EEG preserves the marginal distributions of real EEG features. Figure~\ref{fig:3} presents overlaid histograms comparing real and synthetic EEG for representative spectral and affective features under baseline $z$-score normalization, including Theta, Alpha, and Engagement.

    Across all examined features, GAN-generated EEG closely matches the real EEG distributions in both shape and spread. Real and synthetic data exhibit similar unimodal structure and comparable tail behavior. While small shifts in mean and variance are observed for certain features, these differences are minor relative to the natural inter-participant variability.
    
    Table~\ref{tab:feature_stats} quantitatively summarizes the mean and standard deviation for each feature in both real and GAN-generated EEG. The close agreement in first- and second-order statistics indicates that the GAN captures population-level feature distributions without evidence of overfitting or memorization. These results demonstrate that GAN-generated EEG preserves the marginal distributional structure of real EEG features, providing a necessary foundation for subsequent transition-based modeling. This alignment ensures that synthetic data can serve as a valid input for SBP-based analysis of cognitive state dynamics.
    
    \begin{table}[t]
    \centering
    \caption{Mean and standard deviation of real and GAN-generated EEG features under baseline $z$-score normalization.}
    \label{tab:feature_stats}
    \resizebox{\columnwidth}{!}{%
    \begin{tabular}{lcccc}
    \hline
    \textbf{Feature} & \textbf{Real Mean} & \textbf{Real Std} & \textbf{GAN Mean} & \textbf{GAN Std} \\
    \hline
    Theta       & 0.0797  & 1.0866 & 0.1750  & 0.9550 \\
    Alpha       & 0.0412  & 0.9363 & 0.3523  & 0.8225 \\
    Engagement  & -0.0627 & 1.0493 & -0.5231 & 1.0640 \\
    \hline
    \end{tabular}}
    \end{table}
    
    \subsection{Schrödinger Bridge Energy on Real EEG}
    
    We then apply the Schrödinger Bridge framework to real EEG data to quantify cognitive transport costs between task conditions. SBP energy is computed for transitions from baseline (P1) to subsequent task portions (P2 and P3) for each participant.
    
    Across participants, SBP energy generally increased from the P1 $\rightarrow$ P2 transition to the P1 $\rightarrow$ P3 transition, although Participants 7 and 8 showed decreases and Participant 9 remained approximately unchanged. Overall, this pattern suggests increased cognitive effort with task progression for most participants. Although absolute energy values vary between individuals, the direction of change is largely consistent, indicating that SBP energy captures task-dependent cognitive structure in real EEG.
    
    Participant-level SBP energies for real EEG are summarized in Table~\ref{tab:sbp_real}.
    \begin{table}[t]
    \centering
    \caption{Participant-level SBP energies for real EEG across task transitions.}
    \label{tab:sbp_real}
    \resizebox{\columnwidth}{!}{%
    \begin{tabular}{lcc}
    \hline
    \textbf{Participant} & \textbf{P1 $\rightarrow$ P2 Energy} & \textbf{P1 $\rightarrow$ P3 Energy} \\
    \hline
    Participant 1    & 0.000391 & 0.000491 \\
    Participant 2   & 0.000518 & 0.000549 \\
    Participant 3    & 0.000725 & 0.000740 \\
    Participant 4   & 0.000692 & 0.000907 \\
    Participant 5   & 0.000790 & 0.001063 \\
    Participant 6  & 0.000508 & 0.000835 \\
    Participant 7   & 0.000686 & 0.000556 \\
    Participant 8  & 0.000544 & 0.000443 \\
    Participant 9     & 0.000695 & 0.000694 \\
    Participant 10     & 0.000858 & 0.002588 \\
    \hline
    \end{tabular}}
    \end{table}
    
    \subsection{Schrödinger Bridge Energy on GAN-Generated EEG}
    
    To evaluate whether synthetic EEG preserves transport geometry, we compute SBP energy using GAN-generated EEG under identical conditions. Table~\ref{tab:sbp_gan} reports SBP energies for GAN EEG across the same task transitions.
    
    At the group level, GAN EEG SBP energies are comparable in magnitude to those computed from real EEG. Mean transport costs for P1 $\rightarrow$ P2 and P1 $\rightarrow$ P3 transitions fall within the same order of magnitude, with overlapping variability. This suggests that GAN-generated EEG preserves the overall geometry of task-dependent cognitive transitions. These findings demonstrate that GAN-generated EEG preserves the transition geometry captured by SBP-derived transport costs. This confirms that synthetic EEG retains the dynamical structure required for energy-based modeling of cognitive state transitions, enabling its use in downstream neuroadaptive system applications.
    
    \begin{table}[t]
    \centering
    \caption{Participant-level SBP energies for GAN-generated EEG across task transitions.}
    \label{tab:sbp_gan}
    \resizebox{\columnwidth}{!}{%
    \begin{tabular}{lcc}
    \hline
    \textbf{Participant} & \textbf{P1 $\rightarrow$ P2 Energy} & \textbf{P1 $\rightarrow$ P3 Energy}\\
    \hline
    Participant 1    & 0.000845 & 0.000885\\
    Participant 2   & 0.000860 & 0.001127\\
    Participant 3    & 0.000558 & 0.000951\\
    Participant 4   & 0.000753 & 0.000932\\
    Participant 5   & 0.000581 & 0.000830\\
    Participant 6  & 0.000951 & 0.000948\\
    Participant 7   & 0.000693 & 0.000741\\
    Participant 8  & 0.001016 & 0.000836\\
    Participant 9     & 0.000740 & 0.000830\\
    Participant 10     & 0.000768 & 0.000779\\
    \hline
    \end{tabular}}
    \end{table}

    \subsection{Group-Level Comparison of Real and GAN EEG Transport Costs}
    
\begin{figure}[htbp]
    	\centering
    	\includegraphics[width=0.95\columnwidth]{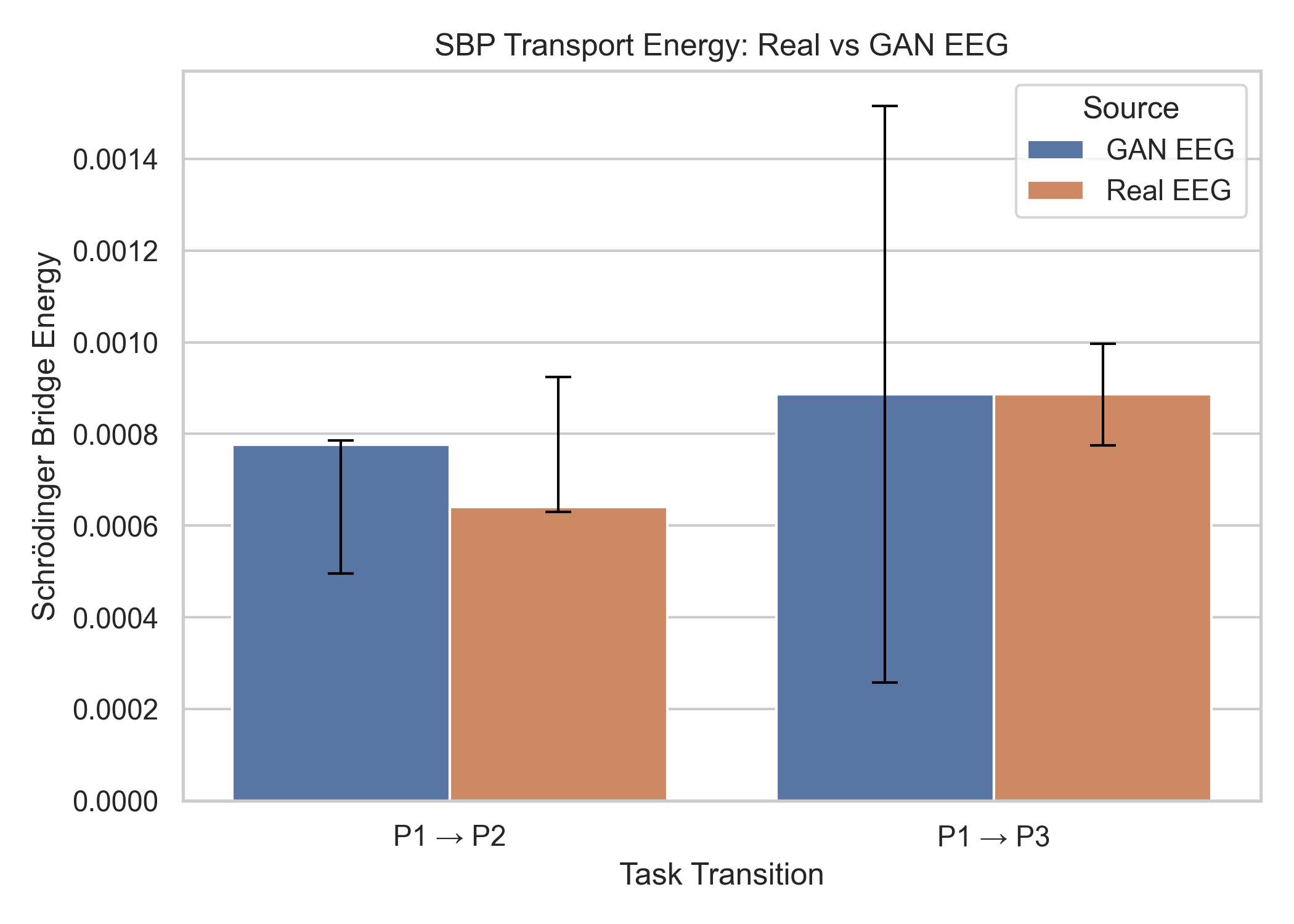}
    	\caption{Group-level comparison of SBP transport energy (mean $\pm$ standard deviation) for real and GAN EEG.}
    	\label{fig:sbp_energy_comparison.png}
    \end{figure}

    To directly compare real and synthetic EEG, SBP energies are aggregated across participants and summarized by task transition. Figure~\ref{fig:sbp_energy_comparison.png} presents the mean SBP transport energy with standard deviation for real and GAN EEG for P1 $\rightarrow$ P2 and P1 $\rightarrow$ P3 transitions.
    
    For both transitions, GAN EEG exhibits transport energies comparable to real EEG. While GAN EEG shows slightly higher mean SBP energy for the P1 $\rightarrow$ P2 transition, the difference lies within one standard deviation and does not indicate systematic distortion of transport structure.
    
    These results demonstrate that GAN-generated EEG preserves group-level cognitive transport costs as measured by SBP energy. This confirms that synthetic EEG retains the transition geometry necessary for energy-based modeling, enabling its use in data-efficient neuroadaptive systems.

    A paired comparison was also performed between real and GAN-generated SBP energies across participants. No statistically significant difference was observed for either the $P1 \rightarrow P2$ transition ($t(9)=-1.61$, $p=0.143$) or the $P1 \rightarrow P3$ transition ($t(9)=0.003$, $p=0.997$). These results indicate that the group-level mean transport energies did not differ significantly between the real and synthetic EEG datasets.

    \subsection{Preservation of Participant-Level Transport Trends}
    
\begin{figure}[htbp]
    	\centering
    	\includegraphics[width=\columnwidth]{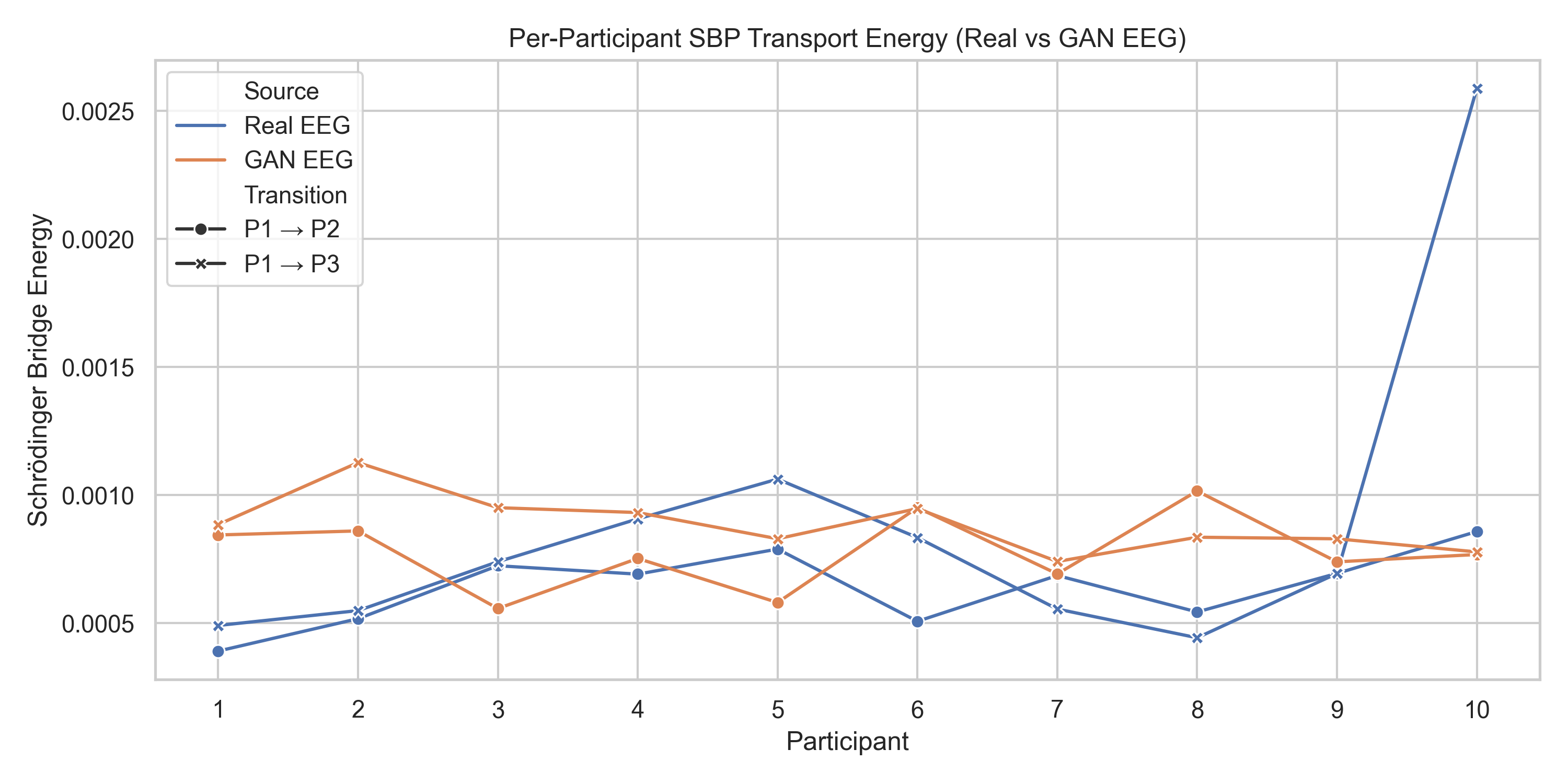}
    	\caption{Participant-level SBP transport energy trends for real and GAN EEG across task transitions.}
    	\label{fig:sbp_energy_per_participant.png}
    \end{figure}

    Beyond group averages, it is critical that synthetic EEG preserves individual-level structure. Figure~\ref{fig:sbp_energy_per_participant.png} shows per-participant SBP energy trends for real and GAN EEG across task transitions. 
    
    The direction of the SBP energy change from the $P1 \rightarrow P2$ transition to the $P1 \rightarrow P3$ transition was preserved for 7 of the 10 participants. Participants 6, 7, and 9 showed mismatched trends, and the magnitude of the participant-level energies differed in several cases, most notably for Participant 10. These results suggest that GAN-generated EEG partially preserves participant-level transition trends, although individual-specific transport geometry is not reproduced consistently across all participants.

\section{NEUROADAPTIVE SYSTEM INTEGRATION}

The proposed framework can be embedded into a closed-loop neuroadaptive human–machine system in which EEG-derived cognitive energy serves as a control signal for real-time adaptation. In this setting, EEG signals are continuously acquired from the user and transformed into baseline-referenced feature representations. These features are used to construct brain-state distributions, which are then evaluated using the Schrödinger Bridge Problem (SBP) to estimate the cognitive energy required for transitions between states. This energy signal provides a quantitative measure of cognitive effort that can be used to guide adaptive system behavior.

\begin{figure}[!t]
    \centering
    \includegraphics[width=\columnwidth]{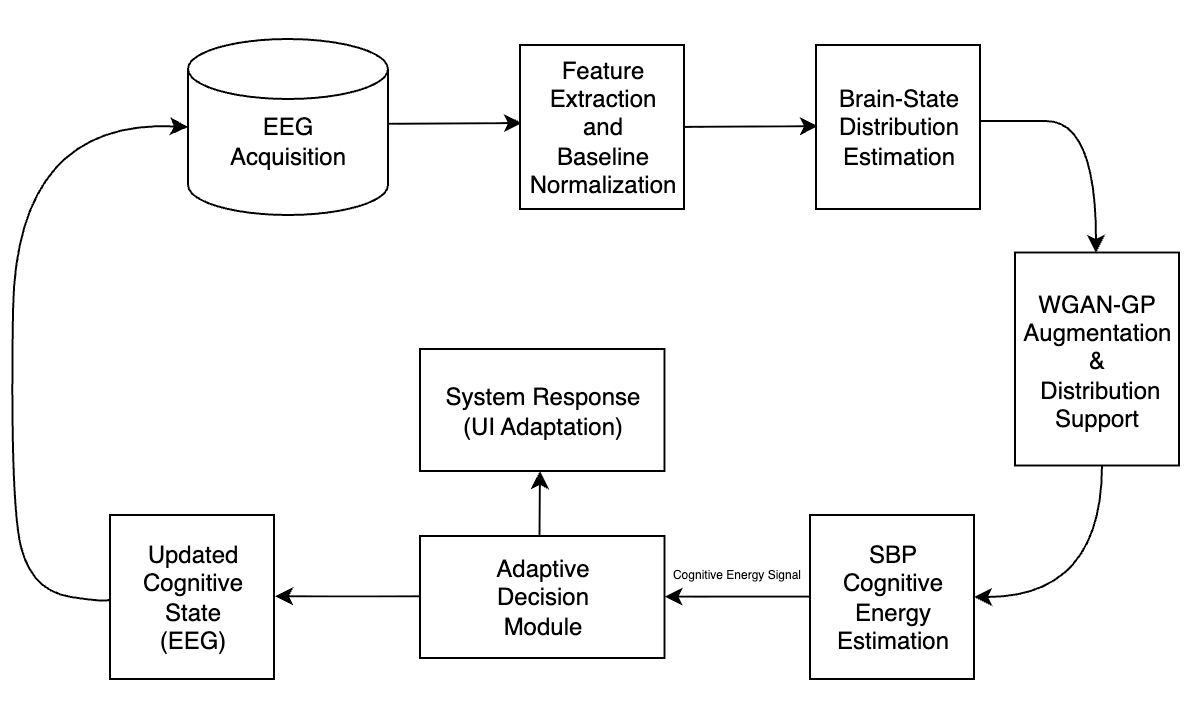}
    \caption{Closed-loop neuroadaptive system framework using EEG-derived cognitive energy as a control signal for real-time human–machine adaptation.}
    \label{fig:system_integration}
\end{figure}

As illustrated in Fig.~\ref{fig:system_integration}, the system operates by first acquiring EEG signals and extracting normalized feature representations that account for subject-specific baselines. These features are used to estimate distributions over brain states, which are supported and augmented using WGAN-GP to improve robustness in data-limited settings. The SBP framework is then applied to compute cognitive transport cost, producing a scalar cognitive energy signal. This signal is passed to an adaptive decision module that determines appropriate system responses based on the estimated cognitive state. The resulting system behavior influences the user’s cognitive state, completing a feedback loop that enables continuous adaptation.

This formulation enables a range of neuroadaptive applications in which system behavior is dynamically adjusted in response to user state. For example, in interactive environments such as games or training systems, elevated cognitive energy may indicate increased mental effort and trigger reduced difficulty or slower pacing, while low energy may indicate under-stimulation and prompt increased challenge. Importantly, because synthetic EEG is shown to preserve SBP-derived transport structure, the proposed framework supports scalable training and deployment of neuroadaptive systems even in settings where real EEG data is limited. This positions cognitive energy as a practical and interpretable control variable for human-centered adaptive systems.
\section{CONCLUSION}

This work demonstrates that GAN-generated EEG preserves the transition geometry required for energy-based modeling of cognitive state dynamics. By using the Schrödinger Bridge Problem (SBP) as a principled measure of cognitive transport cost, we showed that relative energy trends across Stroop task conditions are consistent between real and synthetic EEG at both group and participant levels. These results validate synthetic EEG as a reliable substitute for real data in modeling cognitive state transitions.

Beyond validation, this work establishes SBP-derived cognitive energy as a practical control signal for neuroadaptive human–machine systems. By integrating generative modeling with energy-based analysis, the proposed framework enables data-efficient modeling of cognitive dynamics and supports real-time adaptation of system behavior in response to user state.

This positions synthetic EEG and SBP-based energy modeling as a foundation for scalable neuroadaptive systems, enabling dynamic, user-centered interaction in applications such as adaptive gaming, training environments, and human-centered interfaces.

\bibliographystyle{IEEEtran}
\bibliography{cite}

\end{document}